\documentclass{article}

\newcommand{\comment}[1]{}

\usepackage{amssymb}  

\usepackage{color}
\definecolor{orange}{RGB}{255,127,0}
\definecolor{brown}{RGB}{150,70,0}
\definecolor{green}{RGB}{127,255,127}
\definecolor{darkgreen}{RGB}{0,127,0}
\definecolor{blue}{RGB}{127,127,255}
\definecolor{lightblue}{RGB}{150,150,255}
\definecolor{darkblue}{RGB}{0,0,127}
\definecolor{red}{RGB}{255,90,90}
\definecolor{grey}{RGB}{127,127,127}
\definecolor{pink}{RGB}{255,180,180}

\usepackage{graphicx}

\newcommand{\arxiv}[1]{{{#1}}}

\title{On the universality of cognitive tests  
}
\author
{
	David L. Dowe\\
  {\normalsize\em Computer Science and Software Engineering} \\
	{\normalsize\em Clayton School of Information Technology} \\
  {\normalsize\em Monash University, Vic. 3800, Australia}\\
  {\normalsize \tt david.dowe@infotech.monash.edu.au}\\
$\:$\\
	Jos\'{e} Hern\'{a}ndez-Orallo\\
	{\normalsize\em DSIC, Universitat Polit\`ecnica de Val\`encia, Spain}\\
	{\normalsize \tt jorallo@dsic.upv.es}
}
\date{}

\begin{document}

\maketitle

{
\abstract 
The analysis of the adaptive behaviour of many different kinds of systems such as humans, animals and machines, requires more general ways of assessing their cognitive abilities. This need is strengthened by increasingly more tasks being analysed for and completed by a wider diversity of systems, including swarms and hybrids. The notion of universal test has recently emerged in the context of machine intelligence evaluation as a way to define and use the same cognitive test for a variety of systems, using some principled tasks and adapting the interface to each particular subject. 
However, how far can universal tests be taken? This paper analyses this question in terms of subjects, environments, space-time resolution, rewards and interfaces. This leads to a number of findings, insights and caveats, according to several levels where universal tests may be progressively more difficult to conceive, implement and administer. One of the most significant contributions is given by the realisation that more universal tests are defined as maximisations of less universal tests for a variety of configurations. This means that universal tests must be necessarily adaptive. 
}

\section{Introduction}

The prominent physicist, Stephen W. Hawking, has long
lost his ability to speak, but is relatively fortunate in having
assistants who can follow his facial cues and in having
speech-generation equipment.  We are not aware of any
deficiency in his input (hearing) senses, but we should not
deny his notable intelligence merely on the grounds of his
limited output (limited in both mode and speed) and how
that would severely hinder him on conventional intelligence
tests. Other cases, such as deaf-blind people and those enduring 
locked-in syndrome, are stunning examples of how an individual's 
will to communicate can counteract many
limitations of a hampered (or awkward) communication channel, through
the development of new semantic signals and languages.
Establishing communication through an unconventional channel can
be a real challenge for a peer in those cases, and
may take months, or years, to make the first semantic contact, 
with the subject(s) being considered `vegetative' until that moment.

Another case is that of locked-in syndrome, a condition
which Kate Allatt once had.  She can now communicate
normally, but she had been assumed to be in a vegetative
state until it was realised that she could communicate
---and spell out words--- by blinking her eyes.
A third case in the point is that of Canadian Scott Routley,
again assumed to have been in a vegetative state until it
was very recently discovered by examining MRI brain
scans that he is able to hear and respond to ``yes"/``no"
questions via mental imagery.
These cases describe people whose (auditory) input (or
hearing) appears unaffected but whose intelligence would
not be apparent if we did not allow for their slow output
speed and their less than conventional modes of output.

It is now commonly agreed that intelligence is not specific to (some) humans. More and more life forms are attributed to also have some kinds or degrees of intelligence. The number and ranges of species {\em claimed} to be in this category have been increasing in the past decades, from mammals to molluscs, from swarms to plants \cite{trewavas2005plant}. Nonetheless, there is still an important debate on whether some of them just exhibit a very complex behaviour (evolutionary species adaptation) rather than a truly intelligent behaviour (individual, non-hardwired adaptation). This already complicated picture of intelligent forms is completed (or will be completed) by machines and robots of many kinds, some of them already flaunting some kinds of intelligent behaviour. Similarly, there is a recurrent debate on whether the behaviour is authentically intelligent or just pre-programmed. And, finally, to really make the picture full, we should also consider hybrids (e.g., people with pen and paper, with electronic devices and Internet connection, etc.) and communities (either homogeneous or heterogeneous), where the notions of intelligence and mind may diverge.

Intelligence (or more generally, cognitive) tests are the scientific instruments to measure and categorise the kinds and degrees of intelligence (or more generally, cognitive abilities) of this variety of subjects. Comparative psychology has used many different types of tests to detect and compare abilities across species.
Many tests are particular for some species (or even for some subpopulations or ages in a given species), and only a few have been used for many species, being very specific in terms of tasks or evaluated abilities. One of the most difficult issues is the configuration of an appropriate {\em interface} for each species or individual, in order to guarantee an appropriate administration of the tests, trying to ensure that the subject is focussing on solving a given task.

When moving to the area of artificial intelligence, the use (or just the proposal) of the same test for natural and artificial subjects has been uncommon, apart from Turing's imitation game \cite{turing1950}, and the occasional application of IQ tests to specifically-devoted machines \cite{Evans1963,evans1964,Evans1965,iq,BringsjordSchimanski2003,bringsjord2011}. It is only very recently that the construction of non-anthropocentric intelligence tests for machines has received more attention \cite{Dowe-Hajek1997a,Dowe-Hajek1997b,HernandezOrallo-MinayaCollado1998,Dowe-Hajek1998,mahoney1999text,HernandezOrallo2000a,HernandezOrallo00b,HernandezOrallo00d,Legg-Hutter2007,HernandezOrallo-Dowe2010,AGI2011Evaluating,AGI2011Compression,AGI2011DarwinWallace,CAEPIA2011,LeggVeness2011,Manchester2012}.

In this context, \cite{HernandezOrallo-Dowe2010} introduced the notion of `universal (intelligence) test', as a test that could be administered to any kind of subject, at any speed, and interrupted anytime. The notion of a `test for all' soon generated widespread interest \cite{TheEconomist2011,NewScientist2011}, even though the proposal highlights some concerns about the difficulty (or even the feasibility) of such an idea, especially if we consider the evolution of comparative psychology in the past century. In fact, a new discipline called {\em universal psychometrics} \cite{upsychometrics,upsychometrics2} is suggested as an umbrella area of convergence for the general evaluation of cognitive abilities of any kind of subject (with or without universal tests).

\arxiv{The notion of universal tests does not, in principle, entail what these tests look like and how they are generated. 
Nonetheless, it has been recently argued \cite{HernandezOrallo-Dowe2010,upsychometrics,upsychometrics2}  that the approach for universal tests must be based on computation and, more specifically, on algorithmic information theory (or Solomonoff-Kolmogorov complexity) and (two-part) compression \cite{Solomonoff1964,Wallace-Boulton1968,Wallace-Dowe1999a}. This contrasts with the idea of applying anthropocentric tests for machines, such as the use of the Turing test or common IQ tests to evaluate artificial intelligence. While the reasons for the inadequacy of the Turing Test are easier to understand and have been spotted by many (see, e.g., \cite{HernandezOrallo2000a}), the justification of why IQ tests are not for machines (or for biological systems other than humans) is more cumbersome \cite{IQnotformachines}.}

While most of the above work has been motivated by the lack of proper tools to evaluate machine intelligence (and other cognitive abilities), the same rationale and principles could be used for natural subjects.  
In particular, in this work we analyse the notion of universal intelligence test in a more abstract way and analyse how universal an intelligence test can be. In particular, we will try to answer several questions: Can a test cover any kind of natural or computational subject, operating at any signal resolution and time rates? What resolution aspects should we consider? How crucial is the interface in a universal test? How can we dare measure the intelligence of subjects that we struggle to detect? Do we need intelligence to detect and measure some other intelligent subjects? How are validity, reliability and efficiency affected by a generalisation of a test?

In what follows we will go through these questions with a most general perspective, by considering that subjects can be machines, humans, non-human animals, plants, and other natural, artificial or hybrid systems. We start in section \ref{background} with an overview of the wide range of intelligence tests in natural systems and machines. Section \ref{sec:interfaces} describes what a universal test is and how it depends on the use of different interfaces for different kinds of subjects. Section \ref{sec:resolution} further generalises this idea by discussing possible configurations of time and resolution, and sketches a general procedure for an adaptive universal test. Section \ref{sec:contact} examines more extreme cases where we do not know the rewarding system or we do not even recognise (in the beginning) the agent we want to evaluate. Finally, section \ref{sec:discussion} introduces a hierarchy of test universality levels according to several factors. We discuss the difficulty of universal tests according to this hierarchy and their feasibility for more constrained (but easier) applications that are expected to be common in the near future.

\section{Background}\label{background}

Intelligence \cite{Sternberg2000}, in particular, and cognition \cite{Shettleworth2010}, in more general terms, is nowadays associated with a diversity of species in the natural world. This diversity goes far beyond humans and the animal kingdom, as intelligence has also been recognised (or claimed) in swarms, plants, fungi, immune systems, bacteria, genome and metabolic systems\footnote{The kind of intelligence which is not associated with a centralised brain is sometimes known as non-neural.}  \cite{trewavas2005plant}. Some of the arguments in favour of such a diversity of systems showing some kind of intelligent behaviour are based on examples of classical conditioning \cite{haney1969classical} or some kinds of complex behaviour \cite{applewhite1975learning}, even though some of these findings and claims have also been disputed \cite{sanberg1976neural}. The controversy is stronger for the possibility and detection of extraterrestrial intelligence, and the possibility and construction of machine intelligence. Nevertheless, the terms artificial intelligence and machine cognition are used nowadays to describe systems which lack intelligence and possibly any authentic (embodied) cognition.

Most of this controversy is originated by the lack of agreement of what intelligence is and how it should be measured. One important issue to understand this is the so-called intelligence anthropocentrism, where the very notion of intelligence (and cognition) is frequently said to be a human construct aimed at humans. Psychometrics (see, e.g., \cite{Sternberg2000,Boorsboom2005}) is the most mature (and now robust) discipline in terms of intelligence evaluation, but it is also the most anthropocentric one. As for the measurement of intelligence and other cognitive abilities, one remarkable characteristic in psychometrics is that we can have different tests for the same ability, depending on the kind of subject. For instance, we have different tests for children, disabled people, etc. It is important to note that while the tests (the instruments) are different, the ultimate purpose is to measure the same ability. And here the problems arise, because it is easy to raise objections about whether a test corresponds to an ability\footnote{This objection is often responded with statements such as ``intelligence is what the intelligence tests measure'' \cite{Boring1923}.}. It is also a source of controversy whether  a minor detail in how the test is conceived (or just administered) could have important effects on the measured ability. Nonetheless, on occasions, the use of different tests and interfaces (e.g., a written or an oral test) for different subjects (e.g., a deaf person or a blind person, respectively) is the only way to measure the same ability.

When looking at ways of evaluating cognitive abilities in a general way, comparative psychology (and cognition) \cite{Shettleworth2010,shettleworth2013fundamentals} is the place to look at. First, it has made a very important impact by looking at the problem in a less anthropocentric way. Second, it usually deals with the problem of using different tests for different species in order to measure the same ability. Third, it is an excellent source of how interfaces can be designed, from individuals to swarms. However, its comparative character casts a shadow of relativity in the tests, as the correspondence between the ability and the tests is usually more empirical than theoretical.

A very different approach is found in artificial intelligence. Over the past decades, there have been many proposals to evaluate AI artefacts or areas of the discipline: specific tests (Turing Test \cite{turing1950,oppydowe2011}, the total Turing Test \cite{schweizer1998truly} and other sensorimotor variations \cite{Neumann-etal09}, the Bot Prize \cite{hingston2010new}, CAPTCHAs \cite{von2004telling}, the machine intelligence quotient \cite{zadeh1976miq,bien2002machine}, ... ), competitions (RL competition \cite{whiteson2010reinforcement}, Robocup \cite{kitano1997robocup}, general game playing competition \cite{genesereth2005general}, planning competition \cite{Planningcompetition}, ...) and landmarks (Deep blue  \cite{deepblue2002}, Watson \cite{ferrucci2010building}). 

Of course, one can argue that different tests are required to measure different abilities (e.g., planning, visual recognition, natural language, etc.), but is it reasonable to have different tests for the same ability (e.g., intelligence or learning ability) if the subjects are different? 
As a response to this we should mention that there have been some advocates for the use of the same tests for (at least) humans and machines, namely the use of IQ tests for machines. The roots can be traced to the development and understanding of cognitive models, such as the works of Evans \cite{Evans1963,evans1964,Evans1965} and, indirectly, those of Simon and Kotovsky \cite{SimonKotovsky1963,KotovskySimon1990}. Nowadays, there is a field known as ``psychometric AI'' ---not to be confused with universal psychometrics \cite{upsychometrics,upsychometrics2}---, where IQ tests are used to improve and evaluate artificial intelligence systems. Nonetheless, the most explicit vindication of the use of IQ tests for machines has been recently made by Detterman, editor of the Intelligence Journal, as a response \cite{Detterman2011} to specific domain tests and landmarks (such as Watson). In other words, this view claims that human-level intelligence should be measured with human-level intelligence tests, i.e., IQ tests.

 However, there are many objections to the use of IQ tests for non-human agents, not only for animals (see, e.g, \cite[sec.1.1.4]{Shettleworth2010}) but most especially for machines (see \cite{IQnotformachines} for a full discussion). The main argument is that it is possible to find or construct non-intelligent agents that can score well on classical IQ tests, as has been demonstrated with very small programs (in 2003, Sanghi and Dowe implemented a small program in Perl which could score relatively well on many IQ tests \cite{iq}). This raises serious doubts about the real implications of any non-human (natural or artificial) system that is evaluated using IQ tests (e.g., \cite{eliasmith2012large,yong2012}). It is not clear either that a generalisation or revision of current IQ tests to make them more general is a sound pathway, as we could end up with tests that work with a specific population or in terms of a current technology, but need to be updated recurrently as the population is enlarged and AI technology improves (as happens with the notion of Machine Intelligence Quotient \cite{zadeh1976miq,bien2002machine} or, more blatantly, with CAPTCHAs \cite{von2004telling}).

Despite the variety of tests for humans, non-human biological systems and machines, there is a thing in common for all these tests: they lack a formal (and in most cases, principled) notion of intelligence from which the tests are derived. In other words, the ability is finally defined from what the test measures instead of deriving a test from a principled definition of the ability. At most, for the tests discussed so far, there may be a conceptual analysis of the nuts and bolts of the ability prior to devising the task(s) that may correspond to the ability. However, even in these cases, the link is usually made at an non-mathematical level. Also, the derivation of task difficulty is not obtained from a mathematical definition of the ability but as a refinement of the experimental results of a given population. In the end, this is the same way that thermometers and other measuring devices were designed in the past, without precisely knowing the exact definition of the magnitude to be measured and the physical processes involved.

As an alternative to these approaches, in the past two decades, there have been several efforts to derive definitions and tests of intelligence from computational principles. 
The relevance of (algorithmic) information theory (AIT), or Kolmogorov complexity, to this goal was first hinted at by Chaitin \cite{Chaitin1982} before being independently elaborated upon in \cite{Dowe-Hajek1997a,Dowe-Hajek1997b,HernandezOrallo-MinayaCollado1998,Dowe-Hajek1998,HernandezOrallo2000a,HernandezOrallo00b,HernandezOrallo00d} 
with a series of tests 
noting the relevance and importance of the notions of
(algorithmic) information theory (or Solomonoff-Kolmogorov
complexity) and (two-part) compression
\cite{Solomonoff1964,Wallace-Boulton1968,Wallace-Dowe1999a}. 
An example of one of these tests formally derived from computational principles is shown in Figure \ref{fig:Ctest}, which resembles some exercises found in IQ tests but with a principled generation and assessment of difficulty.

\begin{figure}
\centering
{\sffamily\small

$$
\begin{array}{llll}
k=9  & : & $a, d, g, j, ...$ & $Answer: m$ \\
k=12 & : &  $a, a, z, c, y, e, x, ...$ & $Answer: g$ \\
k=14 & : & $c, a, b, d, b, c, c, e, c, d, ...$ & $Answer: d$  \\
\end{array}
$$
}
\vspace{-0.5cm}
\caption{Examples of series of $Kt$ complexity 9, 12, and 14 used in the $C$-test \cite{HernandezOrallo2000a}.}
\label{fig:Ctest}
\end{figure}

These works were followed some years later by Legg \& Hutter
\cite{Legg-Hutter2007}, who built upon re-inforcement learning
by also using AIT, putting a
Solomonoff-weighted prior distribution over single-agent
environments. A measure (or definition,
but not a test) of intelligence could be theoretically obtained
(in the limit) by seeing what score the agent would obtain
after infinite time in each of the infinitely many environments. 
There are several issues about the feasibility and exact interpretation of such a measure, as raised by \cite{Hibbard2009,HernandezOrallo-Dowe2010}, among others.


\section{Interfaces}\label{sec:interfaces}

From the previous series of contributions on definitions, measures and tests based on AIT, the project {\sffamily anYnt}\footnote{http://users.dsic.upv.es/proy/anynt/} was set to analyse the possibility of constructing the first universal, formal, but at the same time practical, intelligence test. The term universal had been used and understood in different ways by many of the previous proposals, frequently in terms of concepts such as {\em universal} Turing machine, or the use of the term `universal distribution' for any Solomonoff-weighted prior distribution. In this project, however, the term `universal' had a more common meaning and referred to the applicability of the test to any kind of individual, i.e., a test for all.

Pursuing this universality, taking into account the limitations of \cite{Legg-Hutter2007} and other previous approaches, \cite{HernandezOrallo-Dowe2010} introduced an adaptive test which was
{\em anytime} (able to be interrupted at anytime giving a more accurate result as more time is given) and also (supposedly) {\em universal} ---i.e., applicable to machines, humans and non-human animals alike.
While also based on algorithmic information theory, there are some distinctive features of this test, which distinguishes it from \cite{Legg-Hutter2007}. First, the class of environments is carefully selected to be discriminative. Second, while environments are randomly sampled from that class using an a priori distribution (starting with very simple environments), the complexity of the environments adapts to the subject's performance. Third, the test also considers time and, similarly, the speed of interaction adapts to the subject's performance. Finally, the result is not an average of results in all possible environments but an aggregation of how far an agent can reach in terms of the complexity and speed of the environment. 


While some of these features are related, we are mostly concerned in this paper about the alluded universality of the test. If feasible, this universality should allow the comparison of very different subjects with the same tests. As a first proof-of-concept, \cite{AGI2011Evaluating,CAEPIA2011} attempted  an implementation of the test using the environment class introduced in \cite{HernandezOrallo10b}. Actually, humans and AI agents (using some classical reinforcement learning algorithms) were evaluated using the same test. While the exercises were exactly the same, the interface was adapted to each kind of agent. The general idea of a test for all was illustrated, but the results did not show a clear victory of humans over AI agents. There are several possible explanations for this \cite{AGI2011Compression,AGI2011DarwinWallace,Manchester2012,AGI2012social,AISB-AICAP2012a,AISB-AICAP2012b}: it was a prototype, it was not adaptive as the original proposal \cite{HernandezOrallo-Dowe2010}, there was no noise, patterns had low complexity, the environment class was quite limited, no social behaviour or other factors were evaluated, to name a few. In any case, the results do not indicate that the prototype is not a universal test, but rather that it does not measure intelligence properly.

\begin{figure}[ht]
\vspace{-0.3cm}
\centering
\includegraphics[width=0.6\textwidth]{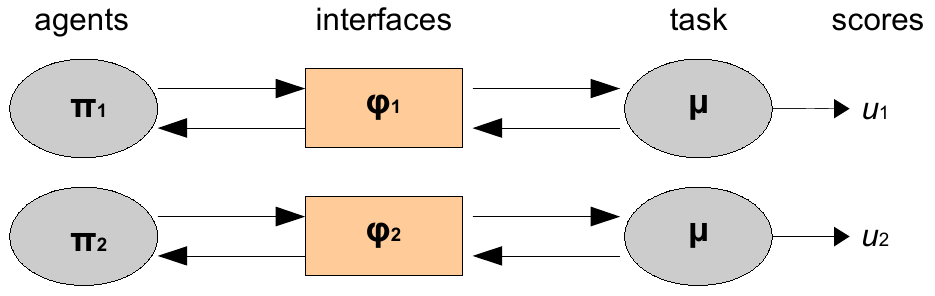}
\vspace{-0.2cm}
\caption{Two interfaces $\phi_1$ and $\phi_2$ for the same task $\mu$ in order to evaluate two different kinds of agents $\pi_1$ and $\pi_2$.}
\label{fig:interfaces}
\vspace{-0.3cm}
\end{figure}

In fact, the main feature of a universal test is that the task must be exactly the same while the interface has to be customised for each individual. In other words, a universal test is composed of a task and an infinite number of possible interfaces. Figure \ref{fig:interfaces} shows this view of interfaces and tasks.

The application of a single task to different species is an everyday chore in comparative psychology, as is the choice of interfaces that are able to engage an animal to do a task (typically by the use of rewards) without adulterating the task itself. However, when the number of species for a single test is limited, the task and the interface are sometimes blurred. 

The meaning and emphasis of a universal (cognitive) test is the intention to apply the task to any possible agent: including humans, non-human animals, plants, communities, hybrids, ..., and most especially machines. Even for those agents that clearly lack the ability (e.g., a flea solving sudokus), the intention is to figure out a way to measure the ability and give a score. This most general way forces us to place the conception of a test in terms of a general, abstract and {\em computational} task. We start with an information-processing ability, define it properly as a set (or distribution) of tasks with a scoring system. Only after this, we may start thinking about interfaces to apply the task to any possible system. This computational approach is now also present in comparative and animal cognition, and particularly in animal evolutionary linguistics, constructing tasks about the kind of grammars (regular, context-free or contextual) several species are able to recognise \cite{gentner2006recursive,pullum2006animal,hauser2007evolutionary,evans2009myth,pullum2009universals}. From there, the evaluators figure out the best way of turning these tasks into tests by the use of appropriate interfaces.

This first understanding of what a universal test is and its possible application to any natural or artificial system also suggests that the question of whether universal tests are possible or not could be further refined into a more informative question: which tasks can be measured with a universal test and those that cannot. In principle, we may be able to construct universal tests for all information-processing tasks but, at the other extreme, it may also be the case that no universal test can be constructed for any task.

Before addressing this question (or the general question of this paper), we need to look at the interfaces more closely.
One reasonable criterion for a fair interface is that it cannot add new information or hide existing information about the task. For instance, if one interface discloses part of the solution while another does not, then the test would be adulterated. One possible mathematical way of defining an interface is by the notion of bijection, as done in \cite{upsychometrics,upsychometrics2}. Nonetheless, there might be computational costs associated with the bijection, so the mapping must have some other constraints.

\begin{figure}[ht]
\centering
\hspace{1cm}\includegraphics[width=0.35\textwidth]{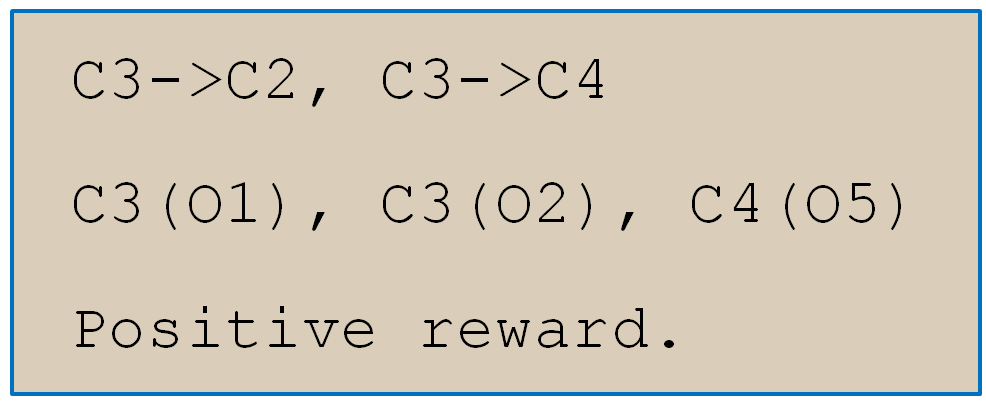}\hfill
\includegraphics[width=0.35\textwidth]{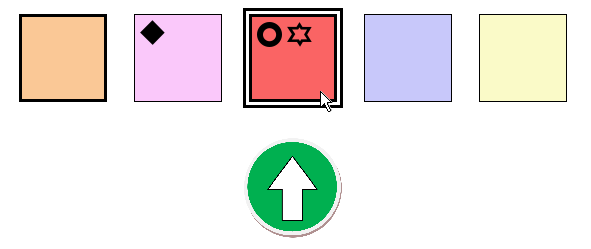}\hspace{1cm}
\caption{Two possible interfaces for the test in \cite{AGI2011Evaluating}. Left: A very raw interface as a character string. Right: A snapshot of an exercise using a graphical interface for humans.}
\label{fig:snapshot}
\end{figure}

Even under these constraints, there are still many possible interface choices for a given agent. One typical approach is to find the best possible interface that does not add any additional information. For instance, humans and animals get much better scores when observations are represented in terms of environments they are used to, i.e., species or culture-friendly interfaces. For instance, Figure \ref{fig:snapshot} shows two different interfaces for the test developed in \cite{AGI2011Evaluating,CAEPIA2011,AISB-AICAP2012a}. While the information is the same and the (computational) effort to transform the observation from one to the other is straightforward, humans would typically score much better with the interface on the right.

Clearly, this search for the optimal interface for each species (or individual) can be applied to any natural and artificial agent.
In artificial agents, exercises are usually presented in terms of data structures (arrays, sets, etc.) instead of visual interfaces, except for those tasks which are precisely evaluating visual recognition or related abilities. 
For natural agents, things are becoming more and more complicated. The interfaces used for some animal species are extremely elaborate and, on many occasions, different interfaces are used for different individuals, according to the understanding and knowledge of the subject's behaviour (e.g., in zoos, aquariums and natural parks, local curators are usually interviewed in order to decide which kind of interaction and rewards may be more appropriate for each individual). In the case of plants or other natural systems, the interfaces are certainly original. In fact, these interfaces are frequently the key to discover abilities in these systems \cite{trewavas2005plant}\cite{haney1969classical}\cite{applewhite1975learning}\cite{sanberg1976neural} that were considered non-existent only a few years ago\footnote{Some of these ideas are being used in better interfaces across species, but also in a pursue of better interfaces between computers and animals \cite{mancini2011animal}.}. 
Humans are not an exception here either, as a great amount of imagination is needed to devise some tests for disabled people, especially for deaf-blinded, using tactile interfaces \cite{arnold2002tactile}  or other approaches \cite{mar1996psychological,vernon1979psychological,ronnberg2001review}.

In general, the interface must pay attention to the sensorimotor characteristics of each agent, including other milieux, such as chemical sensors in animals and plants, and different kinds of data transformation for machines. Many failures in the past have been caused by important mismatches and misconceptions of how animals (and computers) should interact, with a strong anthropocentric bias in our interfaces\footnote{The problem also happens in the other direction: evaluating humans with bit sequences is like evaluating (some) machines or animals with anthropocentric environments.}


This idea of devising a task and looking for optimal environments according to the knowledge that we have about each species and individual has been very fruitful, but it is painstaking in terms of the effort and time that is required to evaluate new species and life forms. Furthermore, this methodology faces more problems with machines, as we can imagine any computable behaviour and we can think of many different kinds of perception, as more and more sophisticated sensor devices are conceived.

So the grand question about the notion of an intelligent test is whether a test is able to evaluate a completely unknown agent. The general idea is that we can only assume some minimal information about the environment or {\em milieu} (either physical or virtual) where the agent is placed. From here, a real universal test should try to find the best interface (signals, time rate, resolution, rewards) which leads to the best score, without providing additional information about the task at hand. 
As a result, a totally universal, adaptive test requires a search in this vast area of time rates and resolutions. In the end, this is what animal cognition has done in the past decades in a manual (scientific-oriented) way. Is it possible to envisage such a process in an automated way, at least in some restricted contexts and for some abilities? In other words, if we are given an environment with some agents, is it possible to have a test that tries to find the interface to better evaluate the agents' ability? This is what we explore next.

\section{Time, resolution and universality through test adaptation}\label{sec:resolution}

In comparative cognition, interfaces are usually associated with physical things: a cage with a small door, a peanut as reward, a set of cups, a touch screen, a light bulb, a set of ropes, etc. Thinking of a test that is able to adapt to all these possible physical configurations (and do this automatically) is far beyond reach. 
Instead of this, in this section, we will consider variations of the same physical (or virtual) `milieu'. For instance, given a screen we can think about many possible resolutions and colours, given an audio signal in a range of frequencies we can consider all the possible variations there. In fact, many human tests are still administered with a sheet of paper, and many different interfaces are still possible with this `rudimentary' milieu.

If we focus on cognitive abilities as information-processing tasks we can fix the milieu and examine the possible variations around it. With this restriction, any interface is in the end a pair of input and output communication channels (in terms of information theory) with a given bandwidth. If we consider discrete interfaces, we can describe this in terms of the input/output resolution and a refresh rate\footnote{Although the terms are exactly the same for screens, we consider any possible milieu here, either auditive, visual or other.}. 
For instance, considering time, if the task is the addition of two natural numbers lower than 10 represented in a unary system, and we agree on the representation of the numbers in the output channel, there is still the question of how much time the numbers are going to be displayed and how much time the agent is going to be allowed to give an answer. If we fix these values we make the evaluation possible for some agents but this also exclude others. For instance, plants are now claimed to do some kind of cognition \cite{garzon2011plants}, but their time-scale is much slower than those of animals.

The anytime test introduced in \cite{HernandezOrallo-Dowe2010} arguably addresses part of the issue of time adaptively by starting with a very fast interaction rate and slowing it down as the results from the agent are not good. The direction of the time change (from very fast interaction to slower interaction) is reasonable as many agents would also react appropriately if the interaction is neither too fast nor too slow, and starting at fast interactions makes the adaptation feasible in finite time. While this is a first approach for making a universal test adaptive on time, there are more issues in terms of time-scale than those reflected by \cite{HernandezOrallo-Dowe2010}. Also, other kinds of resolutions are not considered.

If we take a closer look at time rate, we see at least two time frames that could be (adaptively) increased/decreased:

\begin{itemize}
\item Working time, which can be the time between questions and answers in a questionnaire-like test or it can be the time other agents (e.g., predators) take to make actions or the time the environment makes rewards available. This is in fact what \cite{HernandezOrallo-Dowe2010} adapts. This time typically includes the time the agent needs to act (or write an answer).
\item Exposition time: the amount of time the agent gets to be given the information before it is removed\footnote{As an example of humans requiring more time than some animals, Ayumu the chimpanzee \cite{inoue2007working} is able to note the locations of the numbers 1 to 9 (of which there are 9! = 362880 possibilities) after only 60 milliseconds observation time. A large proportion of humans struggle just to see (and recall) the location of even one of these numbers in this time-frame. This also happens with other milieux, e.g., auditive signals, as the case of a non-native speaker requiring slower speech or repetitions.}. This time frame is usually neglected unless this is the goal of a study on short-memory (or other kinds of memory abilities, such as photographic memory). However, if the exposition time frame is not appropriate, the agent may completely overlook some important data of a task.
\end{itemize}

\noindent On occasions we want a cognitive ability to consider time, such as measuring `reaction time' \cite{mccarthy1981metric}. 
But if time is not part of the ability (e.g., we may want to know the ability to sort a series of numbers, without considering speed) then we (or the test) need to find the optimal time windows for working and exposition time in order to make the test feasible. Let us use the term {\em time configuration} for the set of parameters for a given working and exposition time.

\begin{figure}[ht]
\centering
\includegraphics[width=0.4\textwidth]{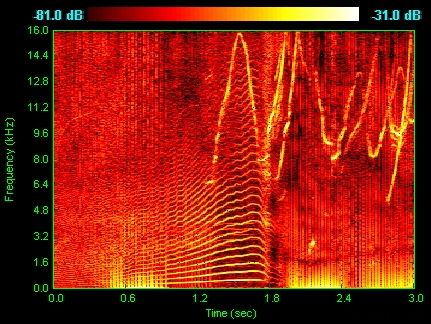}
\caption{Dolphins perform a sophisticated use of the sound spectrum at different resolutions. Whistles and whines are used for communication and can be seen as horizontal striations and vertical lines, respectively. Clicks are used for echolocation, which are visible as strong, big wedges on the top-right of the spectrogram. (GNU-licensed image from wikimedia: http://commons.wikimedia.org/wiki/File:Dolphin1.jpg).}
\label{fig:dolphin}
\end{figure}

A related, but different thing, is resolution. Although we typically think in terms of spatial resolution, it is very useful for the discussion that follows to think of a case where we consider an audio signal, where resolutions appear on the same signal, using, e.g., frequency and amplitude. For instance, figure \ref{fig:dolphin}  \cite{wilson2007intense,madsen2012dolphin} shows how a sound signal can carry many different types of information at several resolutions.

Not only the resolution may be too coarse or too detailed for the agent to see any relevant pattern, but it can take infinitely many representations. Note that the detection of the appropriate resolution is different from any pattern that the signal may carry. This distinction is important, even though both things (resolution and pattern) are usually closely intertwined\footnote{The recognition of patterns depends on a correct resolution configuration. This is crucial to the notion of emergence, and can be translated to communication and interaction as well, since structures emerge from some low-level constituents that have no meaning by themselves. This recognition of patterns at a previously unknown resolution and time, and its relation with emergence, has also been explored (see, e.g., \cite{flecker2011partial}), and can also be referred to the cognitive structures and constructs the agent is creating all along its life \cite{HernandezOrallo00c}.}. Nonetheless, while animals (including humans) usually have innate preprocessing systems that may be used to capture some resolutions (and ignore others) and see patterns in them, it is possible to disentangle one thing from the other. For instance, many artificial pattern (image, speech, etc.) recognition systems have preprocessing devices that render the information ready (e.g., a bitmap or spectrogram) for the analysis of patterns. Working with several resolutions and representations in order to find the most appropriate one may take important time overloads to process. This of course makes resolution and time also closely intertwined in real systems\footnote{It is relevant to mention here that there are some recognition tasks (recognising distorted letters, as in CAPTCHAs) that do not correlate with some other (higher-level) abilities using those letters. In fact, the ability of recognising distorted letters is used as a CAPTCHA because machines are not able to do this well with current technology, not because distorted character recognition is a sign of intelligence.}.

The complexity of resolution and the appropriateness of representation is one of the major issues in artificial intelligence, pattern recognition and machine learning. This issue, however, has been neglected by most machine intelligence tests that we reviewed in section \ref{background}. 
Being conscious of these limitations, let us move forward by considering a communication channel for which we can define any possible resolution, each of them denoted by the term {\em resolution configuration}.

From here, we just define a configuration $\theta$ as a pair of time configuration and resolution configuration. 
We want to evaluate a cognitive ability defined as a distribution of tasks $M$, i.e., a task class.
Consider that we have a set (or distribution) of configurations $\Theta$.  Then we can define:

\begin{equation}
U(\pi, M, \Theta) = \max_{\theta \in \Theta} \lim_{\tau \rightarrow \infty} \Upsilon(\pi,M,\theta, \tau) \label{eq:utest}
\end{equation}

\noindent where $\Upsilon(\pi,M,\theta, \tau)$ is any test on a family of tasks that is applicable to an agent $\pi$ during time $\tau$ (for instance, time-bounded adaptations of \cite{Legg-Hutter2007} or some of the non-adaptive versions introduced in \cite{HernandezOrallo-Dowe2010} or implemented in \cite{AGI2011Evaluating}). Eq. \ref{eq:utest} above defines (or generalises) a universal test from a non-universal test.
The expression $\Upsilon(\pi,M,\theta, \tau)$ is an aggregate over the set of tasks $M$. This must be based on the result on single tasks, $\Upsilon(\pi,\mu,\theta, \tau)$ and can be done in many different ways. One possibility is to weight by task probability: $\Upsilon(\pi,M,\theta, \tau) = \Upsilon(\pi,\mu,\theta, \tau) p(\mu)$ (as in \cite{Legg-Hutter2007}), while another possibility is to define task difficulty and get the result in terms of this difficulty, as in \cite{HernandezOrallo-MinayaCollado1998,HernandezOrallo2000a,upsychometrics,upsychometrics2}.

This maximisation can be translated into the goal of finding the configuration such that the test result is optimal for the subject. This leads to adaptive tests, which search for the appropriate configuration, or at least one that gives an approximation of Eq. \ref{eq:utest} above. In the case we want the test to adapt, there is a need to have some interactive feedback from subject to testers, in terms of the score the subject is achieving.
With this we are ready to introduce a first general procedure for an adaptive universal test. 
Figure \ref{fig:adaptive} shows a general procedure for evaluating subject $\pi$ in an adaptive way.

\begin{figure}
\centering
\begin{enumerate}
\item The history $H$ is initially empty.
\item \label{repeat} Choose a task $\mu$ from $M$, a configuration $\theta$ from $\Theta$ and a time slot $\tau$ taking $H$ into account.
\item Evaluate during time $\tau$ and agent $\pi$ with task $\mu$ at configuration $\theta$, and store the result $R$. 
\item Add $\left\langle \tau, \mu, \theta, R \right\rangle$ to the history $H$.
\item Go to \ref{repeat}.
\end{enumerate}
\vspace{-0.5cm}
\caption{A test that adapts to time configuration and resolution configuration.}
\label{fig:adaptive}
\end{figure}

\noindent The relevant part of the previous procedure is how we select tasks and configurations, especially after the first iteration, using the history of results and the tasks and configurations previously used. 
This can be seen as an extension/generalisation of \cite{HernandezOrallo-Dowe2010} (and ultimately of \cite{Legg-Hutter2007} as well, with the appropriate modifications). For instance, if the agent's score has been poor, we can either try to find a simpler task or change the configuration (which may imply a change in the time configuration, the resolution configuration or both). On the contrary, if the agent's score has been good, we would be tempted to keep the configuration and change to a more difficult (or more informative) task.

An important question is how to obtain the final result of the test. Ideally, if we were able to evaluate all possible configurations and all possible tasks for an infinite amount of time $\tau$ each, the result would be calculated by taking the aggregated performance on the task class (using its distribution or the difficulty function) for the configuration that has given the best result. In practice, in finite time, the test should start with configurations not taking too much time (so $\tau$ can be small) and try as many resolutions as possible as the time configuration uses larger slots. 

So, a universal test for an unknown agent would be a test that makes all the possible efforts to find a configuration with the evaluee at the best resolution and time configurations such that the subject can be evaluated in optimal conditions. Not coincidentally, this is what animal cognition usually does when designing an experiment, and, in the most difficult cases, finding the correct configuration may take decades. Also, as we will discuss in section \ref{sec:discussion}, we are never sure that the right configuration has even been found.

These difficulties appear even though we have already considered that the (physical) communication milieu is fixed (e.g., a sound channel or a screen) and we have also assumed that we recognise the agent and know its reward system. 
As mentioned above, we are not going to consider every possible physical milieu. 
Instead, in what follows we will consider an environment including both the evaluator and the evaluee (e.g., the real world, but most especially, virtual words, such as games, social networks, etc., because of the possible applications and the higher feasibility of a test of this kind, see, e.g., \cite{hernandez2013profiles}). 




\section{Making contact: detecting intelligence}\label{sec:contact}

As said above, we now consider that evaluator and evaluee are agents in an environment. We assume that both can have effect on the environment through actions and they can perceive changes in the environment. This does not necessary mean that the environment has a spatial configuration where they can `move', as in \cite{HernandezOrallo10b}, but just that the possibility of interaction exists (e.g., an environment could just be a series of communication channels as in the matching pennies game, the Turing Test or any configuration in between \cite{Manchester2012}). We also consider that the evaluee has a goal or reward system that is determined by the things that happen in the environment. This setting is assumed in multi-agent systems, games and many other virtual, robotic or physical environments, from artificial intelligence to biology.

A first situation we can consider is when we know the evaluee's reward system so that the evaluator can directly use this by giving more or less rewards to the evaluee according to its performance. This sets an important advantage because we can condition the system to do things. Still, we do not know what communication channel or milieu to use, just how to administer rewards. So the big challenge here is to determine the way to communicate in the environment. In order to do this, the evaluator needs to perform actions in order to see some reactions from the evaluee, in order to understand how to `condition' the evaluee (to try) to perform a task. With animals, we typically use food for this (to approach them, make them trust the evaluator, etc.).

Even in a simplified virtual environment, as found in multi-agent systems or multiplayer games, recognising which channel the agents may use is not always easy. In general, there may be many possible channels, and communication can switch to more efficient communication channels with time\footnote{It is common that pre-established communication channels or protocols are omitted and agents end up communicating through other means that were originally not meant for communication (because the new channel is more efficient, can be concealed from other agents, or other reasons).}. The communication channel may be extremely original, as we can see in many examples of intra- or inter-species communication, from primates \cite{zuberbuhler2000interspecies} to bacteria \cite{federle2003interspecies}. Nonetheless, there are many kinds of communication which are not usually  recognised as such explicitly, as the communication that takes place whenever predators and preys meet, where the very movements and peer positions are an authentic exchange of information. In biology, there is a huge variety of ways of contact and communication, and many are still to be unveiled. In virtual or technological environments this diversity is also very large now. 


Performing a universal test under these circumstances would imply a generalisation of the procedure in Figure \ref{fig:adaptive} by including a variety of channels (along with the diversity of configurations and tasks). In order to determine that the channel is working we would still rely on the reward system and start with very easy tasks, in order to raise the probability that the evaluee makes a correct action (possibly by chance) so it can get positive rewards and then start `focussing' on the task.

Still, a much more challenging problem is when we do not know the agent's reward system. 
Without knowing the agent's goals (or rewarding system) we may detect reaction, but it is much more difficult to properly apply a test. 
The most natural option is to try to learn agent's goals or rewards by observing habits, resistance to change, interacting, etc. 
This is again what ethologists can do, or what children do when playing with bugs or other animals in order to understand their behaviour. 
In order to automate this process (at least partially), there are several information-theoretic options that could be used to detect (and spur) agent-environment or agent-agent interaction \cite{tarapore2006quantifying,williams2010information}, but other approaches exist (relational dynamics, information structure, and many others). 
Independently of what technique is finally used (and whether they can work in general or not), if the evaluator is able to discover the agent's goals then the test could start as in the previous case, using a generalisation of Figure \ref{fig:adaptive}. In this case, some information about the interaction that was used to determine the rewards could also be reused as history for the selection of milieu, time configuration and resolution configuration.


Finally, the most challenging situation is when we have not even recognised the agent we want to evaluate. In this case, we are left in an environment (physical, robotic or virtual) and we need to discover whether there are intelligent forms there and try to evaluate them. 
One of the first caveats in this situation is that there may be many agents, and they can work collectively. As a consequence, even if we are able to recognise the individuals, we may fail at properly recognising intelligence if this only appears as an emergent property of the collection of agents and not of the individuals (as in swarm intelligence). The recognition of the goals/rewards of the collective and some way to communicate with it (in a unified way) is usually more difficult than the recognition of more integrated intelligent forms.

\begin{figure}
\centering
\includegraphics[width=0.39\textwidth]{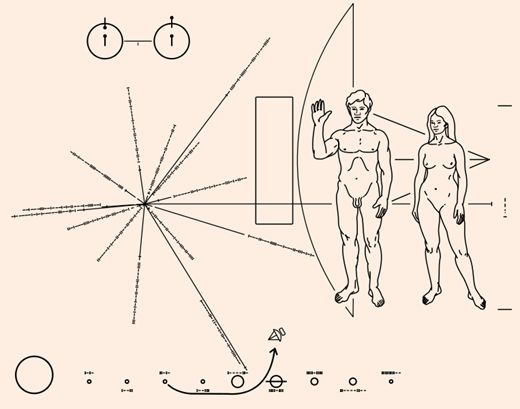}\hfill\includegraphics[width=0.55\textwidth]{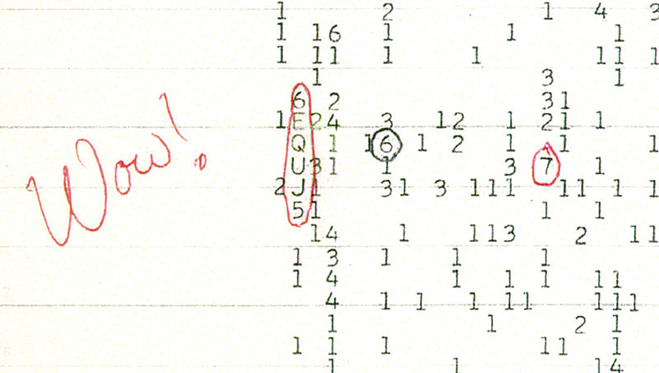}
\caption{Left: Pioneer 10-11 plaques (1972-1973) designed by Frank Drake and Carl Sagan, artwork by Linda Salzman (public domain image from NASA). Right: SETI's Wow! Signal detected by Jerry R. Ehman in 1977 (public domain image from the Ohio State University Radio Observatory and the North American AstroPhysical Observatory).}
\label{fig:contact}
\end{figure}

Either in the form of an individual or a collective, this extremely challenging case has been occasionally discussed in biology but most especially in astro-biology. While astro-biology looks for physical signs of complex biological molecules, the detection of extraterrestrial intelligence through signals from outer space has been subject of much interest in the past decades. However, there is no clear methodology about how to do it (what to scan, as with the SETI project, and what to send, as with the Voyager space probes). Figure \ref{fig:contact} shows an example of an outgoing message (as with the Voyager space probes) and an example of an alluded incoming message. 


One problem is to recognise intelligence from such a transmission (in either way),
and a related secondary problem is to be able to discern and interpret
the message's intended communication. The interpreting of such a message would presumably be best done
by the (Bayesian) algorithmic information-theoretic approaches of
Solomonoff prediction \cite{Solomonoff1964} and/or Minimum
Message Length (MML)
\cite{Wallace-Boulton1968,Wallace-Dowe1999a,Wallace2005}. 
If the (algorithmic) information content (or Solomonoff-Kolmogorov
complexity) of the communication is very low, this suggests something
ultra-regular like a pulsar, analogous to a human repeating the same sound. 
If the Kolmogorov complexity is very high, this suggests unstructured
random background gibberish, perhaps analogous to an incoherent human
infant.  For the message to stand out, it must be like, e.g., (human) language
and have some structure without being totally repetitive.
This should mean having some incrementality in its complexity, with sections of the
message depending on previous sections. 
In terms of what to put in such a message or how to decode such a
message, Wallace (private communication) \cite{Dowe2008a}
considered explaining arithmetic, then eventually Turing machines,
the eventually the Lyman series and Hydrogen, etc.  Solomonoff
likewise wrote much about training sequences\footnote{Although related, we should distinguish the goal of training sequence (turn a non-intelligent, possibly Turing-complete system into an intelligent system (see also \cite{potential,hernandez2013potential}) from the conception of a sequence such that it can be self-understood without common previous knowledge. We will discuss about this issue later on.}
\cite{solomonoff1962training,solomonoff1984perfect,solomonoff2010algorithmic}\cite[sec.7(dolphin talk)]{solomonoff1967inductive}.

This `contact' problem corresponds very neatly to our extreme case of a universal intelligence test where a priori we do not even know about the existence of intelligence forms in an environment and, if they exist, where they are and what they look like.   
One important difference, though, is that in our case we assume the possibility of interaction\footnote{This is also possible in theory with interstellar communication but with a very slow time-frame.}. This means that the principles could be the same, but the degree of repetition in the signals could be highly reduced when the interaction starts. When trying to systematise this, we could use yet another generalisation of the procedure in Figure \ref{fig:adaptive} where the iteration would start looking for agents. The use of information-theoretic tools seems to be appropriate here as well. 
The mere recognition of an agent in these terms is ambitious as an agent can emerge (e.g., by autopoiesis) from very simple rules. For instance, these embodied, minimal cognitive agents have even been found (as gliders) in very minimalistic environments such as Conway's game of life \cite{beer2004autopoiesis}. As there may be many different kinds of agents we may need tools to determine those that are merely interactive, cognitive or finally intelligent. 
Any procedure or test that could be able to eliminate those agents that are not good candidates for further levels would be useful. For instance, once an interactive system is found we could first determine whether we have a cognitive system, adapting some of the tasks of the cognitive decathlon and related approaches \cite{anderson2003newell,mueller2008adapting,mueller2007bica,mueller2008turing,simpson2008refining,langley2011artificial,calvo2003nonclassical,mueller2008adapting} (although many of these approaches are focussed on evaluating cognitive architectures rather than evaluating actual cognitive systems through multi-factorial scores).

\section{Conclusions}\label{sec:discussion}


The difficulty of recognising (and ultimately) measuring intelligence would depend on a number of factors, as shown in Table \ref{tab:levels}. This table shows a gradual view which consists of five possible levels based on four criteria: knowledge of the configuration, milieu, rewards and agents. On occasions we may have some other combinations of these factors or we may have partial information about the configuration (or no information at all about the configuration but still some information about how the reward system works). Also, more information might be available for some agents than others. This suggests the use of Table \ref{tab:levels} as an indication of the factors that need to be taken into account.

\begin{table}
{\center
\begin{tabular}{ccccc}
LEVEL & Configuration & Milieu & Rewards & Agent \\ \hline
1 & $\checkmark$ & $\checkmark$ & $\checkmark$ & $\checkmark$  \\
2 & $\times$ & $\checkmark$ & $\checkmark$ & $\checkmark$  \\
3 & $\times$ & $\times$ & $\checkmark$ & $\checkmark$  \\
4 & $\times$ & $\times$ & $\times$ & $\checkmark$  \\
5 & $\times$ & $\times$ & $\times$ & $\times$  \\
\end{tabular}
\caption{Several levels of universality of a cognitive test depending on the information we have about the time and resolution configuration, the communication milieu, the reward system and the agent itself.}\label{tab:levels}
}\end{table}

The less knowledge we have, the more difficulty the evaluation will be, and we will get lower reliability, efficiency and (probably) validity. 
In addition, there is also another factor that we need to consider: how the universal test is devised, depending on whether the test is conceived in a passive (as classical paper and pencil IQ tests), interactive (as games, computerised adaptive testing, the anytime test \cite{HernandezOrallo-Dowe2010} or any generalisation of figure \ref{fig:adaptive}) or ultimately intelligent way. In this regard, the notion of an `intelligent intelligence test' may sound strange but it is not new, as this is the way intelligence has been detected and evaluated for many centuries, as in interviews and other personal (psychological) assessments. In fact, the Turing test is one example of an `intelligent intelligence test', since itself requires two intelligent subjects to evaluate a third.

It can be argued that the more intelligent the evaluator is the more effective the test can be. While this may be true (especially because a superintelligent being may be able to learn the best procedures to do intelligence testing as they are discovered), it raises many questions about reliability and validity, since an intelligent being may not follow a clear procedure or may not even be fair and objective in the assessment. As there is a subtle line between adaptability and intelligence, we prefer to envisage universal intelligence tests which are highly adaptive but follow a formal and standardised procedure. 

In terms of reliability, we can also see (from the way the procedure in figure \ref{fig:adaptive} works) that even when the proper configuration, milieu, reward system and agent are found then the assessment will still be an {\em under-estimation}\footnote{It is still possible to give over-estimations, such as humans assigning more intelligence to those animals or machines which have an android look, with anthropocentric sensorimotor components, as they look more intelligent than they really are. 
This issue is less frequent and there is always the possibility of double-checking (once the milieu, resolution, reward and agent are fixed).}. This originates from the way the overall result of the test is understood: the procedure tries to find the best conditions for administering the test (i.e., the measure is a maximum). As a result, the evaluation is biassed to under-estimate intelligence (as has happened for many years for non-human intelligence and use to happen with other human cultures). This is a well-known problem in animal cognition (and also happens with other places where intelligence is sought, such as plants, bacteria, SETI, etc.). Moreover, this under-estimation also happens in human psychometrics, where the term ``potential intelligence'' is applied to ``test potential'' \cite{mahrer1958potential,thorp1959predicting,little1972potential}, (not to be confused with the term ``potential intelligence'' applied to the ability of becoming intelligent \cite{potential,hernandez2013potential}).

From a computational point of view, we can say that evaluating intelligence (in a universal context) is, at most, semi-computable\footnote{As we do not consider intelligence a Boolean property but a quantitative one, we do not use the term semi-decidable here.}. This comes from the fact that given an environment that may contain intelligence, there are infinitely many configurations, milieux, reward systems or agents, and there may always be some of them that have not been explored (because of the halting problem). In theory, we could turn this into a computable function or even a function that can be calculated in bounded space and time, by assuming that configurations, milieux, reward systems or agents are resource-bounded.


Finally, there is a risk of over-estimation if the evaluator ends up training the agent instead of evaluating it.
During a very long exploration for optimal configurations we may (inadvertently) train the agent we want to evaluate and make it more intelligent than it was. In fact,  there are training sequences \cite{solomonoff1962training,solomonoff1984perfect,solomonoff2010algorithmic}\cite[sec.7(dolphin talk)]{solomonoff1967inductive} for any universal Turing machine such that the machine becomes intelligent (in fact, as much intelligent as we want, as it can simulate any behaviour, as shown and fully discussed in \cite{potential,hernandez2013potential}). So there is, in principle, a problem in making intelligence tests too long, as the agents can be domesticated (trained) to become intelligent\footnote{This problem can vanish for some artificial agents if they can be reinitialised for each iteration of the adaptive procedure.}. If the agent is trained (or just domesticated), the test mixes actual and potential intelligence (in the sense of \cite{potential,hernandez2013potential}), and the reliability and validity of the test are highly compromised.

%



Let us now summarise the results of our exploration about the limits and caveats of universal intelligence tests. We have seen 
how intricate the notion of universal intelligence test is and how difficult its implementation can be, as far as the challenge is set at the highest level in the hierarchy shown in Table \ref{tab:levels}. Nonetheless, the difficulty also depends on the environments and kinds of agents we want to explore. If we consider all possible natural and artificial agents in our physical world, a fully universal test working in a reasonable amount of time is just a theoretical idea. However, if we consider some simple environments (e.g., cellular automata, multi-agent environments or hybrids \cite{hernandez2013complexity}) or situations where we have part of the information of the hierarchy shown in Table \ref{tab:levels}, we may still establish universal intelligence tests for those niches.

In fact, there are current situations where we need to determine the intelligence of a completely unknown agent. This is becoming more and more common in virtual environments, such as social networks, games, working groups, member authorisation procedures, etc., where we may find agents whose intelligence is completely unknown \cite{hernandez2013profiles}. The Turing Test, custom IQ tests and CAPTCHAs are unsatisfactory in general terms (and will be less so as artificial intelligence advances). As new artificial agents, living organisms, collectives and hybrids thereof become mainstream, we will need to better understand the concept of test universality and devise more effective universal tests for intelligence and other cognitive abilities.

\section*{Acknowledgements}

We would like to thank Paco Calvo for interesting discussions on plant intelligence and other kinds of non-neural intelligence, and for providing us with very useful pointers on this topic.

\bibliography{biblio}

\end{document}